\documentclass[10pt,twocolumn,letterpaper]{article}
\usepackage[pagenumbers]{cvpr}
\usepackage[T1]{fontenc}

\DeclareTextFontCommand{\texttt}{\fontencoding{OT1}\ttfamily}
\usepackage{amsmath,amssymb,booktabs,array,tabularx}
\usepackage{xcolor}
\usepackage{colortbl}
\definecolor{reportblue}{RGB}{38,100,151}
\usepackage[breaklinks,colorlinks,allcolors=reportblue]{hyperref}
\hypersetup{pdftitle={GPT-6-Astra Lights Up Embodied Navigation: Evaluation in Zero-Shot Vision-and-Language Navigation in Continuous Environments},pdfauthor={Guangzhao Dai, Qianru Sun, Qi Wu, and Bin Zhu}}
\newcolumntype{Y}{>{\raggedright\arraybackslash}X}
\newcommand{\astra}{GPT-6-Astra}
\graphicspath{{figures/}}
\title{GPT-6-Astra Lights Up Embodied Navigation\\[3pt]
{\large Evaluation in Zero-Shot Vision-and-Language Navigation in Continuous Environments}}
\author{Guangzhao Dai$^1$,
Qianru Sun$^1$,  
Qi Wu$^2$, 
Bin Zhu$^{1\dagger}$\\
$^1$School of Computing and Information Systems, Singapore Management University \quad \\
$^2$Australia Institute for Machine Learning\\
\normalsize{$^\dagger$Corresponding author and project lead}\\[3pt]
\normalsize{Website: \url{https://daiguangzhao.github.io/gpt-6-astra-for-vln/}}
}
 
\makeatletter
\patchcmd{\@maketitle}{\vskip .375in}{\vskip 0pt}{}{}
\patchcmd{\@maketitle}{\vspace*{12pt}}{\vspace*{4pt}}{}{}
\patchcmd{\@maketitle}{\vspace*{24pt}}{\vspace*{12pt}}{}{}
\g@addto@macro\@maketitle{%
  \vspace{-24pt} 
  \begin{center}
    \includegraphics[width=\textwidth]{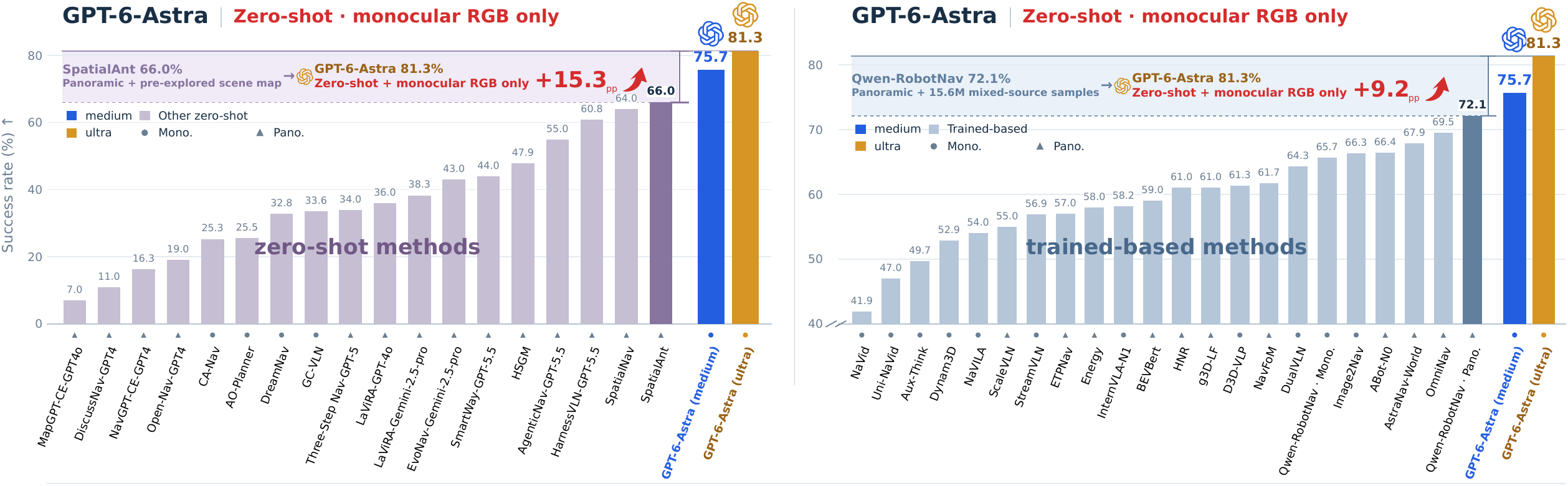}
    \captionsetup{type=figure}
    \vspace{-10pt}
    \caption{\textbf{Strong zero-shot navigation performance of GPT-6-Astra.} With monocular RGB, SR is 75.7\% (medium reasoning) and 81.3\% (ultra reasoning). \astra{} (ultra reasoning) exceeds the best reported zero-shot and trained results by 15.3 and 9.2 percentage points. Following the commonly used zero-shot evaluation protocol~\cite{qiao2025opennav,shi2025smartway,dai2026evonav}, we use the 100-episode R2R-CE subset~\cite{qiao2025opennav} for evaluation. Trained-method results use the full validation-unseen split. Although not strictly fair, these comparisons help the VLN community assess the competitiveness and research value of zero-shot navigation with advanced foundation models.}
    \label{fig:teaser}  
  \end{center}   
  \vspace{3pt} 
} 
\makeatother     
 
\begin{document} 
\maketitle
    
\begin{abstract}
We investigate whether GPT-6-Astra, a general-purpose foundation model, can navigate unfamiliar environments using its own perception, reasoning, and decision-making capabilities.
Our evaluation focuses on zero-shot vision-and-language navigation in continuous environments (VLN-CE) through a minimal interface in the Codex harness, aiming to unleash GPT-6-Astra's full potential for navigation. 
Using monocular RGB, GPT-6-Astra decides when to observe, how to move, and when to stop, without navigation-specific fine-tuning, a trained waypoint predictor, or a pre-built scene map. 
Our evaluation yields four key findings and implications. First, \textbf{\textit{GPT-6-Astra achieves strong zero-shot navigation performance using only monocular RGB observations}}. 
On R2R-CE-100, GPT-6-Astra (ultra reasoning) achieves a success rate of \textbf{\textit{81.3\%}}, exceeding the strongest reported zero-shot and even train-based success rates by \textbf{\textit{15.3}} and \textbf{\textit{9.2}} percentage points, respectively. 
Second, \textbf{\textit{GPT-6-Astra exhibits promising capabilities in interpreting multi-stage instructions, understanding the environment, and adjusting routes}}. 
Third, \textbf{\textit{execution and goal-verification failures persist even with ultra reasoning}}. Fourth, \textbf{\textit{these results motivate combining general-purpose model capabilities with navigation-specific expertise}}. 
Based on these findings, future VLN research should investigate which aspects of instruction interpretation, spatial understanding, and navigation decision-making general-purpose models can handle directly, and where navigation-specific learning can extend their capabilities. 
This includes exploring how spatial representations, navigation experience, and learned control skills can improve progress tracking, error recovery, and goal verification while preserving the flexibility to adjust routes.
\end{abstract}             
            
\section{Introduction} 
\label{sec:introduction}
Can a general-purpose foundation model navigate an unfamiliar environment by directing its own interaction with it? 
Here, we use \astra{} to study this question through Vision-and-Language Navigation (VLN), where an agent follows natural-language instructions to reach a destination in a visual environment~\cite{r2r}. Completing this task requires the agent to follow a language instruction to recognize landmarks, ground spatial relations, track its progress, and choose actions. In continuous environments, it must also coordinate turns and forward movements, inspect uncertain situations, recover from mistakes, and stop at the appropriate location~\cite{vlnce}. VLN therefore tests whether a foundation model can bring perception, reasoning, memory, and action together over an extended interaction.
 
Early VLN research developed navigation policies trained on annotated instructions and trajectories~\cite{r2r,speakerfollower}. With the emergence of models such as GPT-4~\cite{achiam2023gpt}, Gemini-2.5-Pro~\cite{geminiteam2025gemini25}, and Qwen2.5-VL~\cite{bai2025qwen2-5VL}, attention increasingly shifted toward transferring general language and visual capabilities to navigation~\cite{chen2024mapgpt_verified,ding2026lavira_icra}. 
Many approaches adapt these capabilities through navigation-specific training or human-designed workflows for planning, memory, and action selection~\cite{zhang2024navid_rss,wei2026streamvln_icra,qiao2025opennav,shi2025smartway}. 
This report does not introduce a new navigation method. Following the harness-based setup of~\cite{zhou2026embodiedcontrol}, we use the Codex harness\footnotemark{} to run \astra{} with a minimal interface for monocular RGB observations and primitive actions. Here, Codex only sustains the session, while \astra{} makes the navigation decisions. This setting allows us to examine the model's capabilities as expressed through its own interaction with the environment. Our evaluation yields four key findings and implications.
 
\textit{\textbf{(1) \astra{} achieves strong zero-shot VLN performance using only monocular RGB observations.}}
Following prior zero-shot evaluation protocols~\cite{qiao2025opennav,shi2025smartway,dai2026evonav,shi2025fastsmartway}, we evaluate \astra{} on R2R-CE-100\footnotemark{}. \astra{} achieves success rates of \textbf{75.7\%} (medium reasoning) and \textbf{81.3\%} (ultra reasoning), with corresponding SPL scores of \textbf{65.6\%} and \textbf{71.5\%}. As shown in Figure~\ref{fig:teaser}, \astra{} (ultra reasoning) outperforms state-of-the-art VLN methods, with SR gains of \textbf{15.3} percentage points over zero-shot methods and \textbf{9.2} points over trained methods~\cite{zhang2026spatialant_verified,zhang2026qwen}. Although evaluation subsets and configurations differ, these results demonstrate superiority in navigation without additional navigation-specific training or a pre-built scene map.

\textit{\textbf{(2) \astra{} exhibits promising capabilities in interpreting multi-stage instructions, understanding the environment, and adjusting routes.}}
Recorded trajectories show \astra{} identifying instructed landmarks, grounding spatial relations, and following successive turns and room transitions in the specified order. With ultra reasoning, the median nDTW among successful trajectories ranges from \textbf{\textit{90.3\% to 91.0\%}} across runs, indicating close agreement with reference paths. Recorded interactions also show the model taking additional observations, returning to earlier locations, and adjusting its route and movement direction. These results highlight \astra{}'s own capabilities for instruction following and adaptive navigation.

\begingroup
\widowpenalty=10000
\clubpenalties 3 10000 10000 0
\textit{\textbf{(3) Reliable route execution and goal verification remain challenging, even with ultra reasoning effort.}}
Across both reasoning settings, 8.0\% of tasks fail in all evaluations, while 30.0\% have mixed outcomes. Failure cases include substantial route deviations despite plausible landmark matches, ineffective movement, and arrival claims inconsistent with the final position. These results distinguish persistent task failures from run-to-run variation and show that higher reasoning effort does not ensure reliable completion in the evaluated setting.
\par\endgroup

\begingroup
\clubpenalty=10000
\textit{\textbf{(4) These results motivate combining general-purpose model capabilities with navigation-specific expertise.}}
Prior VLN research provides benchmarks and methods for visual grounding, memory, and action selection~\cite{r2r,vlnce,speakerfollower,qiao2025opennav,shi2025smartway}. Our findings motivate examining which navigation capabilities general-purpose models can provide directly and where navigation-specific learning can extend them. Spatial representations, navigation experience, and learned control skills could improve progress tracking, error recovery, and goal verification while preserving route flexibility.
\par\endgroup

\footnotetext[1]{\hypertarget{Hfootnote.1}{}\label{fn:codex-session}The Codex harness maintains a continuous session in which \astra{} itself perceives, reasons, and makes navigation decisions.}

\footnotetext[2]{\hypertarget{Hfootnote.2}{}\label{fn:zero-shot-evaluation}Following established zero-shot evaluation protocols~\cite{qiao2025opennav,shi2025smartway,dai2026evonav}, we use the 100-episode R2R-CE subset introduced by Open-Nav~\cite{qiao2025opennav}; trained-method results use the full validation-unseen split. Although not strictly fair, these comparisons help the VLN community assess the competitiveness and research value of zero-shot navigation with advanced foundation models.}
 
\section{Evaluation Setup}
\label{sec:setup}
\label{sec:formulation}
 
\subsection{Navigation Interface}
\label{sec:navigation-interface}

Given a natural-language instruction, \astra{} navigates an unfamiliar indoor scene through low-level actions in continuous space~\cite{vlnce}. Its visual input consists of $512\!\times\!512$ monocular RGB observations. Scene maps, reference trajectories, goal coordinates, global poses, and depth are unavailable through the navigation interface.

\begin{figure}[t]
\centering
\includegraphics[width=\columnwidth]{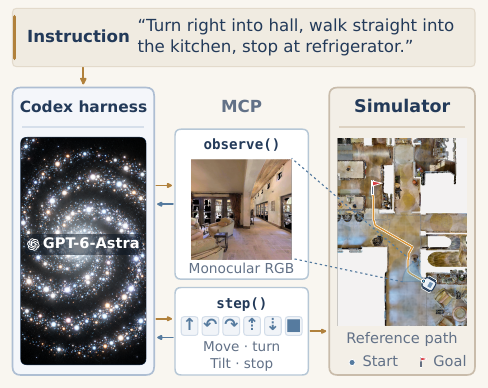}
\caption{\textbf{Minimal navigation interface for GPT-6-Astra.} Within the Codex harness, \astra{} calls \texttt{observe()} and \texttt{step(actions)} via the Model Context Protocol (MCP) to obtain monocular RGB observations and execute primitive actions. This minimal external interface aims to fully unleash the model's own navigation capabilities.}
\label{fig:framework}
\label{fig:navigation-interface}
\end{figure}

Each episode runs within a continuous Codex session. \astra{} itself interprets the instruction and available interaction context to select observations, action sequences, and when to stop. The Model Context Protocol (MCP) interface connects the model to the simulator (Figure~\ref{fig:navigation-interface}). The prompt specifies tool use, budgets, and stopping rules.
 
The \texttt{observe()} tool returns the current camera view without advancing the simulator or consuming an action step. The \texttt{step(actions)} tool executes an ordered sequence of primitive actions: forward motion of 0.25\,m, left/right turns of $15^\circ$, camera tilts of $30^\circ$ up/down, and STOP. It returns execution counts, remaining budget, and termination status; obtaining a new RGB image requires another \texttt{observe()} call.
  
\subsection{Evaluation Protocol and Metrics}
\label{sec:evaluation-protocol}
             
Following prior zero-shot VLN-CE evaluation protocols~\cite{qiao2025opennav,shi2025smartway,shi2025fastsmartway}, we evaluate on the same 100-episode R2R-CE validation-unseen subset.
The subset spans 10 scenes (Footnote~\ref{fn:zero-shot-evaluation}). We use no navigation-specific fine-tuning, trained waypoint predictor, or pre-built scene map.

We evaluate \astra{} with medium and ultra reasoning effort using the same tasks, task prompt, and navigation tools. Each episode has a budget of 500 primitive actions and a 2,400\,s time limit, ending upon STOP or exhaustion of either limit.

Success requires an explicit STOP with a final geodesic distance to the goal below 3\,m. We report success rate (SR) and success weighted by path length (SPL) for task completion and path efficiency. Navigation error (NE), oracle success rate (OSR), and normalized dynamic time warping (nDTW) additionally measure final goal distance, whether any trajectory position enters the success radius, and reference-path agreement, respectively~\cite{vlnce,ilharco2019ndtw_verified}.

\section{Navigation Performance}
\label{sec:results}
\label{sec:sota}

Table~\ref{tab:sota} summarizes navigation performance across learning regimes and observation settings. \astra{} (medium reasoning) achieves 75.7\% SR and 65.6\% SPL, while \astra{} (ultra reasoning) reaches 81.3\% SR and 71.5\% SPL. The differences are 5.7 and 5.9 percentage points, respectively, calculated before rounding. The configuration with ultra reasoning also has higher OSR (83.7\% versus 80.7\%), higher nDTW (74.0\% versus 70.5\%), and slightly lower NE (2.9\,m versus 3.0\,m). These are descriptive comparisons between the evaluated configurations.

\begin{table*}[tp]
\centering
\footnotesize
\setlength{\tabcolsep}{1.5pt}
\renewcommand{\arraystretch}{1.00}
\caption{\textbf{R2R-CE val-unseen results.} Full: complete val unseel split; R2R-CE-100: Sampled 100 episodes by Open-Nav; Author-sampled: authors' samples with unverified Open-Nav correspondence; ``--'': unavailable or unverified. Pano.: $360^\circ$ panoramic observations; Mono.: monocular views. NE: meters; other metrics: percentages. \textbf{Bold}/\underline{underline}: best/second-best. \astra{} results report mean $\pm$ s.d. over three runs per reasoning setting. Text, figures, and rankings use means unless stated otherwise. Evaluation settings vary across methods.}
\label{tab:sota} 
\begin{tabular*}{\textwidth}{@{\extracolsep{\fill}}llccrrrr@{}}
\toprule
Method & Source & Eval. split & \shortstack{View} & NE $\downarrow$ & OSR $\uparrow$ & SR $\uparrow$ & SPL $\uparrow$ \\
\midrule 
\rowcolor{black!6}
\multicolumn{8}{l}{\emph{VLN-CE supervised learning}} \\
ScaleVLN~\cite{wang2023scaling} & ICCV'2023 & Full & Pano. & 4.8 & -- & 55.0 & 51.0 \\
ETPNav~\cite{an2025etpnav_verified} & TPAMI'2025 & Full & Pano. & 4.7 & 65.0 & 57.0 & 49.0 \\
BEVBert~\cite{an2023bevbert} & ICCV'2023 & Full & Pano. & 4.6 & 67.0 & 59.0 & 50.0 \\
HNR~\cite{wang2024hnr_verified} & CVPR'2024 & Full & Pano. & 4.4 & 67.0 & 61.0 & 51.0 \\
Energy~\cite{liu2024energy} & NeurIPS'2024 & Full & Pano. & 4.7 & 65.0 & 58.0 & 50.0 \\
g3D-LF~\cite{wang2025g3dlf_verified} & CVPR'2025 & Full & Pano. & 4.5 & 68.0 & 61.0 & 52.0 \\
NavFoM~\cite{zhang2026navfom_iclr} & ICLR'2026 & Full & Pano. & 4.6 & 72.1 & 61.7 & 55.3 \\
ABot-N0~\cite{chu2026abotn0_verified} & arXiv'2026 & Full & Pano. & 3.8 & 70.8 & 66.4 & 63.9 \\
OmniNav~\cite{xue2026omninav} & ICLR'2026 & Full & Pano. & 3.7 & 74.6 & 69.5 & 66.1 \\
Qwen-RobotNav-8B~\cite{zhang2026qwen} & arXiv'2026 & Full & Pano. & 3.5 & 78.5 & 72.1 & \underline{66.6} \\
AstraNav-World~\cite{chen2025astranav} & arXiv'2025 & Full & Pano. & 3.9 & 73.9 & 67.9 & 65.4 \\
NaVid~\cite{zhang2024navid_rss} & RSS'2024 & Full & Mono. & 5.7 & 49.2 & 41.9 & 36.5 \\
Uni-NaVid~\cite{zhang2025uninavid_rss} & RSS'2025 & Full & Mono. & 5.6 & 53.3 & 47.0 & 42.7 \\
NaVILA~\cite{cheng2025navila_rss} & RSS'2025 & Full & Mono. & 5.2 & 62.5 & 54.0 & 49.0 \\
Aux-Think~\cite{wang2025auxthink_neurips} & NeurIPS'2025 & Full & Mono. & 5.9 & 54.9 & 49.7 & 41.7 \\
Dynam3D~\cite{wang2025dynam3d_neurips} & NeurIPS'2025 & Full & Mono. & 5.3 & 62.1 & 52.9 & 45.7 \\
StreamVLN~\cite{wei2026streamvln_icra} & ICRA'2026 & Full & Mono. & 5.0 & 64.2 & 56.9 & 51.9 \\
DualVLN~\cite{wei2025dualvln} & arXiv'2025 & Full & Mono. & 4.1 & 70.7 & 64.3 & 58.5 \\
InternVLA-N1~\cite{internrobotics2025internvlan1} & Tech.Rep.'2025 & Full & Mono. & 4.8 & 63.3 & 58.2 & 54.0 \\
D3D-VLP~\cite{wang2026d3dvlp_verified} & CVPR'2026 & Full & Mono. & 4.7 & 67.2 & 61.3 & 56.1 \\
Image2Nav~\cite{wang2026image2sim} & arXiv'2026 & Full & Mono. & 4.0 & 72.9 & 66.3 & 61.5 \\
Qwen-RobotNav-8B~\cite{zhang2026qwen} & arXiv'2026 & Full & Mono. & 4.4 & 72.7 & 65.7 & 59.6 \\
\midrule
\rowcolor{black!6}   
\multicolumn{8}{l}{\emph{Zero-shot VLN-CE}} \\
NavGPT-CE-GPT4~\cite{zhou2024navgpt_verified} & AAAI'2024 & Full & Pano. & 8.4 & 26.9 & 16.3 & 10.2 \\
HSGM~\cite{li2026hsgm_verified} & CVPR'2026 & Full & Pano. & 5.4 & 58.7 & 47.9 & 32.8 \\
MapGPT-CE-GPT4o~\cite{chen2024mapgpt_verified} & ACL'2024 & R2R-CE-100 & Pano. & 8.2 & 21.0 & 7.0 & 5.0 \\
DiscussNav-GPT4~\cite{long2024discussnav_verified} & ICRA'2024 & R2R-CE-100 & Pano. & 7.8 & 15.0 & 11.0 & 10.5 \\
Open-Nav-GPT4~\cite{qiao2025opennav} & ICRA'2025 & R2R-CE-100 & Pano. & 6.7 & 23.0 & 19.0 & 16.1 \\
Three-Step Nav-GPT-5~\cite{zheng2026threestepnav} & AISTATS'2026 & R2R-CE-100 & Pano. & 5.9 & 39.0 & 34.0 & 29.1 \\
STRIDER-GPT-4o~\cite{he2025strider} & NeurIPS'2025 & R2R-CE-100 & Pano. & 6.9 & 39.0 & 35.0 & 30.3 \\
LaViRA-GPT-4o~\cite{ding2026lavira_icra} & ICRA'2026 & R2R-CE-100 & Pano. & $6.4\!\pm\!0.28$ & $43.3\!\pm\!3.2$ & $36.0\!\pm\!1.7$ & $28.3\!\pm\!0.8$ \\
LaViRA-Gemini-2.5-pro~\cite{ding2026lavira_icra} & ICRA'2026 & R2R-CE-100 & Pano. & $6.5\!\pm\!0.27$ & $48.7\!\pm\!2.1$ & $38.3\!\pm\!0.6$ & $28.3\!\pm\!0.9$ \\
O2C-Nav-Gemini-2.5-Pro~\cite{pan2026o2cnav} & arXiv'2026 & R2R-CE-100 & Pano. & 5.8 & 65.0 & 49.3 & 30.9 \\
EvoNav-GPT-4o~\cite{dai2026evonav} & CVPR'2026 & R2R-CE-100 & Pano. & 6.0 & 35.0 & 30.0 & 24.9 \\
EvoNav-Gemini-2.5-pro~\cite{dai2026evonav} & CVPR'2026 & R2R-CE-100 & Pano. & 5.0 & 51.0 & 43.0 & 37.8 \\
SmartWay-GPT-4o~\cite{shi2025smartway} & IROS'2025 & R2R-CE-100 & Pano. & 7.0 & 51.0 & 29.0 & 22.5 \\
SmartWay-GPT-5.5~\cite{shi2025smartway} & IROS'2025 & R2R-CE-100 & Pano. & 5.2 & 60.0 & 44.0 & 35.0 \\
AgenticNav-Gemini-2.5-pro~\cite{li2026agenticnav} & arXiv'2026 & R2R-CE-100 & Pano. & 5.9 & 63.0 & 49.0 & 33.2 \\
AgenticNav-GPT-5.5~\cite{li2026agenticnav} & arXiv'2026 & R2R-CE-100 & Pano. & 5.2 & 65.0 & 55.0 & 48.4 \\
SpatialNav~\cite{zhang2026spatialnav_verified} & arXiv'2026 & Author-sampled & Pano. & 5.2 & 66.0 & 64.0 & 51.1 \\
SpatialAnt~\cite{zhang2026spatialant_verified} & arXiv'2026 & Author-sampled & Pano. & 4.4 & 76.0 & 66.0 & 54.4 \\
HarnessVLN-GPT-5.5~\cite{chen2026harnessvln_verified} & arXiv'2026 & -- & Pano. & 4.0 & 72.7 & 60.8 & 43.5 \\
Fast-SmartWay-GPT-4o~\cite{shi2025fastsmartway} & arXiv'2025 & R2R-CE-100 & F3+P & $7.7\!\pm\!0.42$ & -- & $27.8\!\pm\!2.22$ & $25.0\!\pm\!2.70$ \\
CA-Nav~\cite{chen2025canav_verified} & TPAMI'2025 & Full & Mono. & 7.6 & 48.0 & 25.3 & 10.8 \\
AO-Planner~\cite{chen2025aoplanner_verified} & AAAI'2025 & Full & Mono. & 7.0 & 38.3 & 25.5 & 16.6 \\
DreamNav~\cite{wang2025dreamnav} & arXiv'2025 & -- & Mono. & 7.1 & 41.0 & 32.8 & 29.0 \\
GC-VLN~\cite{yin2025gcvln_corl} & CoRL'2025 & Full & Mono. & 7.3 & 41.8 & 33.6 & 16.3 \\
\midrule  
\rowcolor{black!6} 
\textbf{GPT-6-Astra (medium reasoning)} & Report'2026 & R2R-CE-100 & Mono. & $\underline{3.0}\!\pm\!0.4$ & $\underline{80.7}\!\pm\!2.1$ & $\underline{75.7}\!\pm\!1.5$ & $65.6\!\pm\!2.1$ \\
\rowcolor{black!6}
\textbf{GPT-6-Astra (ultra reasoning)} & Report'2026 & R2R-CE-100 & Mono. & $\mathbf{2.9}\!\pm\!0.2$ & $\mathbf{83.7}\!\pm\!1.5$ & $\mathbf{81.3}\!\pm\!2.5$ & $\mathbf{71.5}\!\pm\!1.7$ \\
\bottomrule
\end{tabular*}
\vspace{3pt}
\begin{minipage}{\textwidth}\footnotesize
\textit{Score sources.} Source denotes the method's publication. NaVid, DualVLN, InternVLA-N1, NavFoM, ABot-N0, OmniNav, AstraNav-World, and Qwen-RobotNav scores follow Qwen-RobotNav Table~1~\cite{zhang2026qwen}. MapGPT, DiscussNav, and NavGPT CE scores follow SmartWay, Open-Nav, and CA-Nav, respectively~\cite{shi2025smartway,qiao2025opennav,chen2025canav_verified}. LLM variants follow EvoNav Table~1, SmartWay Table~I, and AgenticNav Table~1; SmartWay-GPT-5.5 is AgenticNav's reproduction~\cite{li2026agenticnav}. LaViRA and Fast-SmartWay report mean $\pm$ s.d. over three and four runs, respectively; O2C-Nav uses main-table means.\par
\textit{Inputs and settings.} F3+P: three frontal views with initial or on-demand $360^\circ$ panoramas. O2C-Nav uses four surrounding views; STRIDER also uses Qwen-VL-Max for perception. SpatialAnt uses scene reconstruction and simulator depth.
\end{minipage} 
\end{table*}

Among zero-shot methods, GC-VLN reports 33.6\% SR and 16.3\% SPL with monocular observations, while SpatialAnt reports 66.0\% SR and 54.4\% SPL with panoramic observations and scene reconstruction~\cite{yin2025gcvln_corl,zhang2026spatialant_verified}. Both \astra{} configurations achieve higher SR and SPL with monocular RGB alone. With ultra reasoning, SR exceeds the best reported zero-shot result by 15.3 percentage points.

Trained methods provide a complementary reference. Image2Nav reports 66.3\% SR and 61.5\% SPL with monocular observations~\cite{wang2026image2sim}; Qwen-RobotNav-8B reaches 72.1\% SR and 66.6\% SPL with panoramic observations~\cite{zhang2026qwen}. Both \astra{} configurations have higher SR, while only the configuration with ultra reasoning exceeds Qwen-RobotNav-8B in SPL. Its SR advantage is 9.2 percentage points. Trained-method results use the full validation-unseen split. Differences in evaluation coverage and system settings make these comparisons contextual, rather than evidence of superiority under matched conditions.

We next examine how \astra{} interprets instructions, uses visual observations, and adjusts its navigation, followed by failures in execution and goal verification.

\raggedbottom
\section{From Instructions to Adaptive Navigation}
\label{sec:analysis}

\begin{figure*}[t]
\centering
\includegraphics[width=\textwidth]{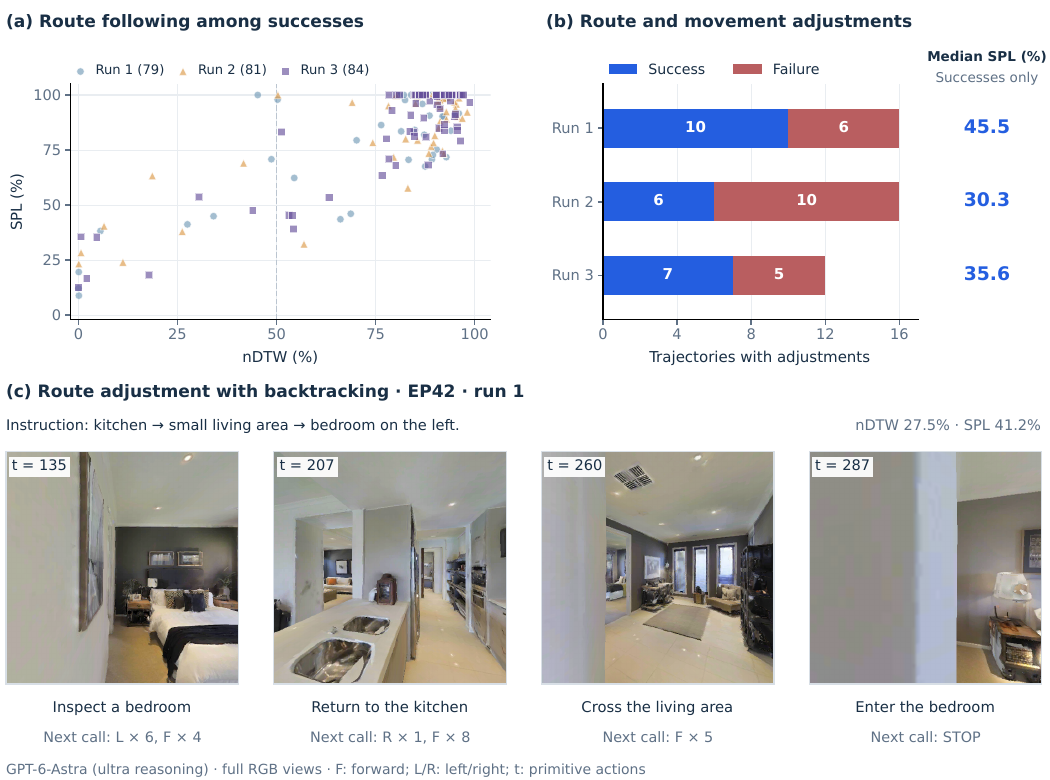}
\caption{\textbf{Route following and navigation adjustments.} \astra{} (ultra reasoning): (a) successful trajectories across three runs; (b) outcomes for trajectories with observed route or movement adjustments, with median SPL among successes in each run; (c) returning to the kitchen before reaching the goal in EP42 (run 1). SPL measures complete trajectories. Intermediate frames are omitted; action labels show the next executed tool call.}
\label{fig:behavior-evidence}
\end{figure*}

We analyze three runs of \astra{} (ultra reasoning) on R2R-CE-100 to examine instruction following, environment understanding, and route adjustment. Navigation metrics summarize task outcomes and reference-path agreement; RGB observations and executed actions show how the model identifies landmarks, follows room transitions, and changes its route or movements. Trajectories from separate runs are repeated observations of the same tasks. Unless stated otherwise, illustrated RGB sequences come from the first run.

\begin{table}[!ht]
\centering
\footnotesize
\setlength{\tabcolsep}{2.4pt}
\caption{\textbf{Navigation performance by instruction type.} \astra{} (ultra reasoning), mean $\pm$ sample SD across three runs. $N$ counts fixed tasks; scores (\%) measure whole-task outcomes. Selection types are defined in Section~\ref{sec:grounding}.}
\label{tab:instruction-diagnostics}
\begin{tabular}{@{}lrrrr@{}}
\toprule
Instruction subset & $N$ & SR & SPL & nDTW \\
\midrule
R2R-CE-100 & 100 & $81.3\!\pm\!2.5$ & $71.5\!\pm\!1.7$ & $74.0\!\pm\!1.2$ \\
Order-based selection & 14 & $95.2\!\pm\!8.2$ & $77.3\!\pm\!8.2$ & $75.9\!\pm\!3.2$ \\
Landmark-based selection & 13 & $74.4\!\pm\!11.8$ & $69.0\!\pm\!8.1$ & $73.9\!\pm\!2.0$ \\
\bottomrule
\end{tabular}
\end{table}

\subsection{How Does GPT-6-Astra Ground Spatial Instructions?}
\label{sec:grounding}

\astra{} can use target order and positions relative to landmarks to guide navigation, with different task-level outcomes across instruction types. We retain two fixed instruction subsets. \emph{Order-based selection} identifies a target by its order or position among alternatives, such as the \emph{second room} or \emph{rightmost doorway}. \emph{Landmark-based selection} identifies a target relative to a reference landmark, using \emph{left/right of}, \emph{between}, or \emph{behind}. Across three runs, mean SR is $95.2\pm8.2$\% for the 14 order-based tasks and $74.4\pm11.8$\% for the 13 landmark-based tasks (Table~\ref{tab:instruction-diagnostics}). These are whole-task navigation scores, not the accuracy of individual spatial decisions. The subsets may overlap and differ in scene layout and route difficulty; score differences therefore do not isolate relation difficulty.

EP218 and EP244 illustrate the behaviors behind these aggregates and succeed in all six evaluations across both reasoning settings. In the first run with ultra reasoning, EP218 follows the hallway past a bathroom identified in the public account as the first room on the left, then approaches the next doorway. In EP244, \astra{} selects the doorway \emph{to the left of the white double doors}. Their final NE values in this run are 2.6\,m and 1.8\,m, respectively. These sequences illustrate spatial selection along successful routes, but do not verify every intermediate interpretation.

\subsection{How Closely Do Successful Trajectories Follow Reference Paths?}
\label{sec:temporal}

Successful trajectories generally show high agreement with reference paths. Across the three runs, median nDTW among successes ranges from 90.3\% to 91.0\%. Figure~\ref{fig:behavior-evidence}a shows all 244 successful trajectories, which are repeated observations of the same task set, not 244 distinct tasks. nDTW measures reference-path agreement rather than the correctness of individual instruction steps.

EP116 illustrates instruction following across successive room transitions. In the first run, \astra{} leaves the closet, crosses the bedroom, and stops in the bathroom beside the tub and sink, with 0.1\,m NE and 89.2\% nDTW.

Some successful trajectories nevertheless deviate substantially from the reference path: 8, 7, and 7 score below 50.0\% nDTW in the three runs, respectively, corresponding to $9.0\pm1.0$\% of successes. EP42 in the first run succeeds with 0.7\,m NE but has 27.5\% nDTW and 41.2\% SPL. Its return to the kitchen illustrates this detour.

\subsection{How Does GPT-6-Astra Adjust Routes and Movements?}
\label{sec:adaptive}

Recorded trajectories show \astra{} returning to earlier locations and changing its route or movement direction during navigation. We count a trajectory when the model reports a route mismatch or difficulty moving forward, changes its subsequent actions, and the corresponding RGB observations support a change in route or approach. Routine orientation and statements without matching execution evidence are excluded. Across the three runs, 16, 16, and 12 trajectories meet this criterion, of which 10, 6, and 7 succeed, respectively (Figure~\ref{fig:behavior-evidence}b).

EP42 illustrates a route adjustment followed by successful completion (Figure~\ref{fig:behavior-evidence}c). After reaching a bedroom, \astra{} checks whether the route matches the instructed sequence, returns to the kitchen, follows its inner aisle, and crosses the small living area into the bedroom on the left. The RGB sequence and executed actions support this return and change of route.

Route adjustment does not ensure an efficient trajectory. Among the successful trajectories counted above, median SPL is 45.5\%, 30.3\%, and 35.6\% in the three runs, respectively, while median nDTW ranges from 8.9\% to 44.3\%. These scores measure complete trajectories, including detours before an adjustment. They characterize selected trajectories with route mismatches or movement difficulties and do not isolate the effect of adjustment.
 
\begingroup
\widowpenalty=10000
\section{Failure Modes and Reliability}
\label{sec:failures}

We next examine where these navigation capabilities fall short of reliable execution and goal verification, and whether failures recur across runs. In the three runs of \astra{} (ultra reasoning), 79, 81, and 84 trajectories succeed. Another 3, 3, and 1 enter the goal radius but fail at termination; 18, 16, and 15 never enter it (Figure~\ref{fig:failure-evidence}a). We examine these outcomes and assess persistent failures using task-level success counts across all six evaluations.

\begin{figure*}[t]
\centering
\includegraphics[width=\textwidth]{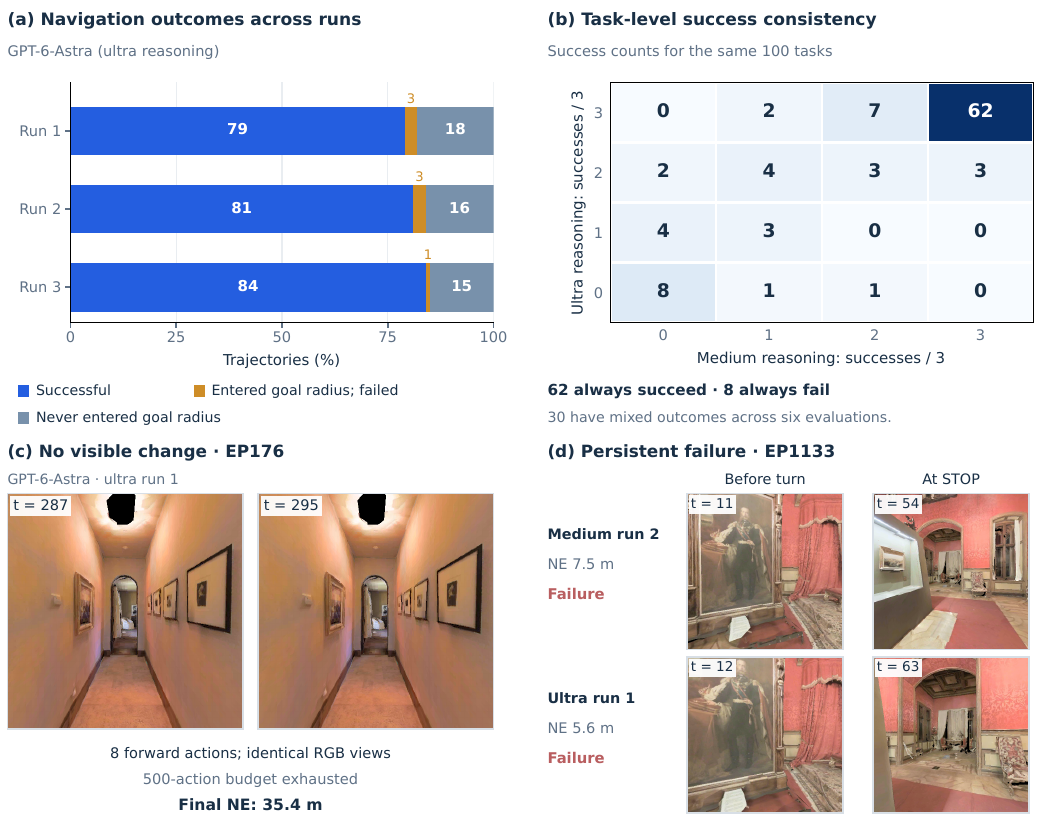}
\caption{\textbf{Repeated evaluations distinguish persistent and variable failures.} (a) Outcomes of \astra{} (ultra reasoning). (b) Each cell counts tasks by successes across three runs per setting. (c) Identical RGB views after eight forward actions in EP176 (ultra run 1). (d) EP1133, which fails in all six evaluations; selected frames are from medium run 2 and ultra run 1. $t$: executed primitive actions.}
\label{fig:failure-evidence}
\end{figure*}

\subsection{Where Does Route Following Break Down?}
\label{sec:route-failures}

Many failed trajectories terminate well beyond the success radius, rather than narrowly missing the endpoint. In each run with ultra reasoning, 10--13 failures end at least 5\,m from the reference goal, including 5--8 more than 10\,m away. Thus, the remaining errors include substantial route deviations across repeated evaluations.

EP513 illustrates how plausible local landmark matches can support an incorrect arrival claim. The instruction specifies a fireplace living area, two white sofas, and the adjacent dining-room entrance. In the first run with ultra reasoning, the final view contains a fireplace and pale seating, and \astra{} reports arrival, yet the final NE is 14.5\,m. The task succeeds in the other two runs with ultra reasoning. This contrast identifies a failure in one observed execution, rather than an inability to complete that task. The recorded views alone do not locate the first route error.

\subsection{Can GPT-6-Astra Recover from Execution Difficulties?}
\label{sec:execution-failures}

\astra{} sometimes resumes progress after ineffective movement, but recovery is not consistent. Across three runs with ultra reasoning, 11 forward-only batches in 10 trajectories have pixel-identical RGB views before and after execution; six trajectories eventually succeed. This analysis covers 1,103 batches with readable observations at both endpoints. Identical views neither establish a collision nor capture all execution difficulties.

In EP70 from the first run, turns and forward movements near a narrow doorway are followed by views from inside the next room, and the task succeeds. In EP176 from the same run, eight forward actions produce identical observations (Figure~\ref{fig:failure-evidence}c). The trajectory later exhausts the 500-action budget with a final NE of 35.4\,m. EP176 fails in all three runs with ultra reasoning, but succeeds in two runs with medium reasoning. The evidence therefore supports inconsistent execution and recovery, rather than an untraversable task or a universal failure under both settings.

\subsection{Does GPT-6-Astra Stop at the Right Place?}
\label{sec:stopping}

A stopping decision does not necessarily verify the instructed destination. \astra{} issues STOP in 96--99 trajectories per run with ultra reasoning, yet each run includes 1--3 failures whose trajectories previously entered the success radius. Reaching the goal vicinity and recognizing a plausible landmark do not ensure a correct stop.

For example, EP705 in the first run with ultra reasoning reports a fire extinguisher beside a doorway and stops with 12.3\,m NE despite OSR equal to one. The same task succeeds in the other two runs with ultra reasoning. This variation motivates checking the final location against the route and instruction rather than relying on a local landmark match alone. OSR records whether the goal radius was entered; it does not establish when this occurred or why the trajectory later ended elsewhere.

\subsection{Does Higher Reasoning Effort Resolve These Failures?}
\label{sec:persistent-failures}

Higher average performance with ultra reasoning coexists with a set of tasks that fail repeatedly under both settings. Figure~\ref{fig:failure-evidence}b counts successes for each task across three runs with medium reasoning and three with ultra reasoning. Sixty-two tasks succeed in all six evaluations, eight fail in all six, and the remaining 30 have mixed outcomes. Of the 14 tasks that fail in every run with medium reasoning, six succeed at least once with ultra reasoning, while eight remain unsuccessful. Conversely, three tasks succeed in all runs with medium reasoning but fail in one run with ultra reasoning. These observations show that increased reasoning effort does not ensure reliable task completion; they do not identify the cause of each persistent failure.

EP1133 is one of the eight tasks that fail in all six evaluations. Its final NE ranges from 7.5 to 11.4\,m with medium reasoning and from 5.6 to 12.3\,m with ultra reasoning. The selected trajectories in Figure~\ref{fig:failure-evidence}d approach a portrait, turn left, and stop facing an arched doorway; both remain outside the 3\,m success radius. A lower NE in one run is therefore not evidence of reliable improvement. Repeated evaluations distinguish persistent errors from variable outcomes more reliably than a single trajectory pair.

\par\endgroup
\section{Implications for Embodied Navigation}
\label{sec:discussion}

\subsection{Do General-Purpose Models Offer Another Route to VLN?}

The present results support another route to VLN through direct interaction with the environment. Within the evaluated interface, \astra{} selects observations, movements, and stopping decisions without additional navigation-specific fine-tuning. Recorded trajectories show it interpreting instructions and adjusting routes during execution. These behaviors motivate examining which capabilities general-purpose models can provide directly and where navigation-specific learning can help.

Prior VLN research provides benchmarks and methods for visual grounding, spatial memory, and action selection~\cite{r2r,vlnce,speakerfollower,qiao2025opennav,shi2025smartway}. How best to divide perception, planning, and control between general-purpose models and specialized methods remains an empirical question.

\subsection{How Can Navigation Expertise Complement General-Purpose Models?}

Spatial representations could help relate current observations to previously visited places and completed instruction steps. A useful representation would preserve landmark relations and route history while allowing the model to revise uncertain spatial judgments. Such representations could support progress tracking and goal verification in visually similar locations.

Navigation experience could support adaptation through learning from trajectories or demonstrations supplied at inference time. Pairing observations and actions with their outcomes may help the model use past interactions when choosing routes or recovering from mistakes. The research question is which experiences remain useful across different instructions and scene layouts.

Learned control skills could provide efficient movement and recovery routines that a general-purpose model can select or revise~\cite{zhang2024navid_rss,wei2026streamvln_icra}. Execution feedback could help detect ineffective motion, while goal verification could compare the current view with the instruction and traversed route. These additions should preserve the model's flexibility to observe and adjust routes.

Our evaluation motivates these directions but does not test the proposed additions. Because added components may introduce errors or restrict useful model behavior, their contributions require controlled evaluation.

\subsection{How Should Future Navigation Systems Be Evaluated?}

Future studies should compare direct model control with variants that add a spatial representation, an experience mechanism, or a learned skill. Comparisons should fix the model, tasks, observations, primitive actions, and budgets, and disclose any added information. Component ablations and repeated runs can establish which failures change and whether gains persist across evaluations.

Evaluation should measure task success, path efficiency, and reference-path agreement alongside progress tracking, recovery, and stopping behavior. It should also test whether an added component preserves the flexibility to acquire new observations and adjust routes. Inference cost and execution latency should be reported alongside navigation outcomes. Tests across environments, instruction lengths, and layouts should assess transfer beyond R2R-CE-100.

\section{Conclusion}
\label{sec:conclusion}
We evaluated \astra{} on R2R-CE-100 through a minimal interface for monocular RGB observations and primitive actions within the Codex harness. SR reaches 75.7\% with medium reasoning and 81.3\% with ultra reasoning, without navigation-specific fine-tuning in our evaluation. Recorded interactions reveal promising capabilities in interpreting multi-stage instructions, understanding the environment, and adjusting routes. Execution and goal-verification failures also persist: eight tasks fail in all six evaluations. These findings motivate studying how navigation-specific learning can extend general-purpose models' spatial understanding and action capabilities while preserving flexible observation and route adjustment.

\noindent\textbf{Limitations and future work.} Our evaluation measures repeatability on a fixed R2R-CE-100 task set; broader generalization remains to be tested. Future evaluations will include additional datasets, such as RxR-CE and NavRAG-CE, as well as more diverse environments and instructions. High inference cost and latency currently constrain practical deployment. Historical reductions in inference cost at comparable capability levels~\cite{maslej2025aiindex} suggest that these barriers may diminish as models and inference systems improve. 
Finally, GPT-6-Astra is closed-source, and its training data are not available to us. We cannot tell whether OpenAI used public or internally collected navigation data, or how much was used. Here, zero-shot means that we did not fine-tune the model for navigation; it does not imply that the model had never seen navigation data during training.
{\small
\bibliographystyle{ieeenat_fullname}
\bibliography{main}
}
\end{document}